## Title

Enhancing the interpretability of spatially variable $N_2O$ model predictions with soft sensors during wastewater treatment.

## Author list

Mohammad Raeisi Gahrouei[1,2], Pedram Ramin[3], Vincenzo A. Riggio[2], Carlos Domingo-Félez[1*]

[1] Infrastructure and Environment, School of Engineering, University of Glasgow, University Avenue, Glasgow G12 8QQ, United Kingdom

[2] DIATI, Politecnico di Torino, C.so Duca Degli Abruzzi, 24, 10129, Torino, Italy

[3] PROSYS, Department of Chemical and Biochemical Engineering, Technical University of Denmark, 2800 Kgs., Lyngby, Denmark

*Corresponding Author: E-mail: Carlos.Domingo-Felez@glasgow.ac.uk, Tel: +44 (0)141 330 4170

**Graphical Abstract**:

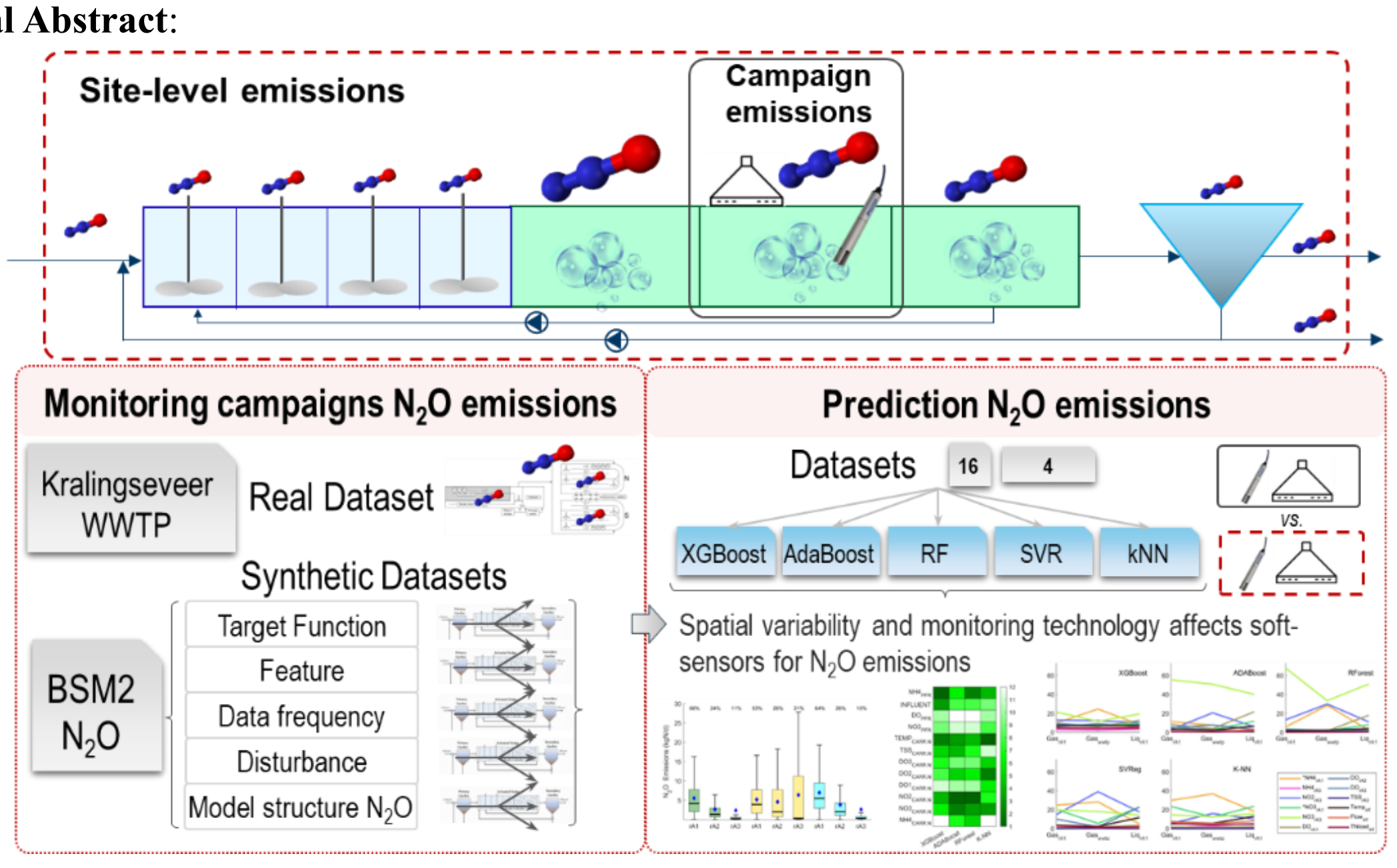

**Abstract**

Model-based solutions for nitrous oxide ($N_2O$) emissions from wastewater treatment plants (WWTP) are informed by operational datasets designed to control nutrient levels in liquid waste, coupled with dedicated campaigns for $N_2O$ measurements. We analysed how machine learning (ML) models predict disturbances to WWT operation and spatially variable $N_2O$ emissions. A real dataset was investigated to validate the modelling framework from $N_2O$ emissions predicted by four ML models ($R^2$ = 0.79 - 0.89). Monitoring campaigns for $N_2O$ were simulated with a plant-wide mechanistic model to include additional sensors, site-level $N_2O$ datasets, and wastewater disturbances (n = 16). ML models were highly accurate (0.97 ± 0.02, n = 80), but the feature importance depended on the model, the scenario and the $N_2O$ measurement scale (reactor *vs.* WWTP). We argue that $N_2O$ soft sensor model predictions are limited to the measuring location and the methodological uncertainty of the dataset, which affect the interpretability of the model. Lastly, the analysis of the mechanistic model structure exposed interactions between autotrophic and heterotrophic pathways over nitric oxide which can overestimate aerobic nitrite production and bias the $N_2O$ pathway contributions.

## 1. Introduction

Nitrous oxide ($N_2O$) is produced biologically as a by-product of nitrification and obligate intermediate of denitrification. $N_2O$ emissions depend on the wastewater treatment (WWT) technology, the operational strategy, the microbial community and climate. Multiple time-scales impact $N_2O$ production and emission simultaneously: microbial reactions, fluid dynamics across the WWTP, diurnal wastewater loadings or climate seasonality. Additionally, the spatial scales also vary across orders of magnitude, from microscopic to macroscopic interactions. Hence, $N_2O$ emissions during WWT vary over time and space across multiple scales, and monitoring a single bioreactor during a weekly campaign does not capture the representative site-scale emissions.

One key challenge for monitoring $N_2O$ emissions is the nature of gaseous emissions, ubiquitous across open surfaces in the WWT plant, but emission rates vary orders of magnitude (Zhao et al., 2025). In comparison, standard methods for liquid streams are established and the methodological uncertainty is lower. Most long-term monitoring campaigns extrapolate point-measurements such as gas hoods (~1m$^2$) or liquid sensors in single locations to unit-scale emissions. The spatial variability of $N_2O$ emissions is challenging to capture experimentally because every monitoring campaign is constrained by the type, the number, and the location of a limiting number of $N_2O$ sensors. To date, a dataset from long-term site-scale measurements with individual contributions of each water treatment processes does not exist. Instead, the focus has been on aerated treatment where biological nitrogen removal seems to contribute the most to total $N_2O$ budgets, as observed by short-term site-scale measurements (Delre et al., 2017). Hence, the Emission Factors (EF, gas emitted / N-load) estimated for site-level calculations are limited by the existing monitoring techniques, i.e. estimations from point-measurements do not account for the spatial variability, and predictions from short-term datasets do not capture the long-term temporal variability. Given the limited availability and reliability of physical $N_2O$ sensors, models for $N_2O$ emissions can act as soft sensors, to generate higher spatial and temporal data, that aid develop mitigation strategies.

Mechanistic models (McM) for $N_2O$ emissions are extensions to models for N-removal that incorporate our understanding of the $N_2O$ production pathways. While no consensus exists on the mathematical description of the biological pathways, McMs have predicted the effects of substrates such as nitrite ($NO_2^-$) or dissolve oxygen (DO) to the pathway contributions successfully in lab-scale environments (Peng et al., 2014). Application to full-scale long-term operation remains challenging but the trends are captured in industrial and municipal treatment (Flores-Alsina et al., 2025; Lei et al., 2025).

Statistical and machine-learning (ML) methods describe $N_2O$ emissions via a black-box approach. ML models predict $N_2O$ emissions from point-measurements accurately, and multiple procedures and frameworks exist that focus on improving data preprocessing techniques or the model calibration and validation steps (Khalil et al., 2023; Vasilaki et al., 2020). The number of alternative methods and degrees of complexity has increased rapidly over the last years from multiple regression to deep learning models (Hwangbo et al., 2021; Vasilaki et al., 2018). Modelling results depend on the target function (i.e. predicted variable) of the model, which depend on the datasets from the monitoring campaign: liquid $N_2O$ concentration during 14 months (Hwangbo et al., 2021), gaseous measurements from one out of seven parallel lanes during 15 months (Seshan et al., 2025), or two gaseous point-measurements during 96h (Szeląg et al., 2023). Regardless of the type of model, WWTP layout, or target functions, ML models are often reported to be highly accurate ($R^2 = 0.94$ (Hwangbo et al., 2021), 0.94 (Khalil et al., 2023), 0.98 (Seshan et al., 2025)). The use of soft sensors is a promising solution to reveal the spatial variation in $N_2O$ emissions.

The accuracy of existing ML models is linked to one dataset for a specific WWT technology, operation, and $N_2O$ measurement. Hence, assessing the robustness and reliability of these ML models seems essential to understand under which conditions models will fail (e.g. alternative $N_2O$ monitoring locations, process disturbances) or the limits to their interpretability. For example, the analytical uncertainty of technical errors, liquid and gaseous measurements, or stripping rate calculations ranges between 10 to 25% (Bollon et al., 2016; Domingo-Félez et al., 2024; Myers et al., 2021), but is seldom considered to detect model underfit or overfit.

While the specific McM structure biases the synthetic datasets, it allows diagnosing potential effects that are currently extremely hard to isolate experimentally. By combining one real full-scale dataset with synthetic, plant-wide datasets generated by a McM we: a) investigate how spatially variable $N_2O$ emissions affect the performance of five ML models, b) assess the performance of ML models against synthetic disturbances of wastewater treatment, c) identify structural limitations of McM commonly used to predict $N_2O$ emissions.

## 2. Methods

### 2.1. $N_2O$ emission datasets

#### 2.1.1. Kralingseveer WWTP.

The Kralingseveer WWTP plant treats domestic wastewater and consists of a plug-flow reactor followed by two parallel carrousel reactors with surface aerators as shown in Fig. 1 (left) and Fig. S1.1. Carrousel reactors were covered and the off-gas analysed for $N_2O$ emissions which reduced potential biases from the spatial variability expected across carrousels due to dissolved oxygen and stripping gradients. More details can be found in the original study (Daelman et al., 2015).

The original dataset contains multiple online and laboratory datasets with non-matching timestamps. Initially, and to allow for comparison of modelling results, the same features were considered as in as (Khalil et al., 2023) except for $NH_4^+$ loading since the flow rate and the influent ammonia concentration are available: 12 non-$N_2O$ features across the WWTP (Influent (x1), Plug Flow Reactor (x3), Carrousel (x8)) and one $N_2O$ feature (Carrousel North). However, the dataset contains additional $N_2O$ measurements from covered tanks: Carrousel South and PFR. We consider the total emissions from Kralingseveer the key target function ($WWTP_{tot}$) because gaseous $N_2O$ emissions occur across the WWTP. Nevertheless, four target functions for $N_2O$ were analysed: Carrousel North ($Carr_N$), sum of Carrousels North and South ($Carr_{tot}$), average of two sampling points in the PFR ($PFR_{tot}$), and $WWTP_{tot}$ (**Fig. 1**, **Table 1**). After data preprocessing the dataset extends for 314 days (March 2011 – January 2012), and the actual duration is 228.84 days with a sampling frequency of 25 minutes (n = 13182). Details on data pre-processing, WWTP layout and target functions (SI-S1).

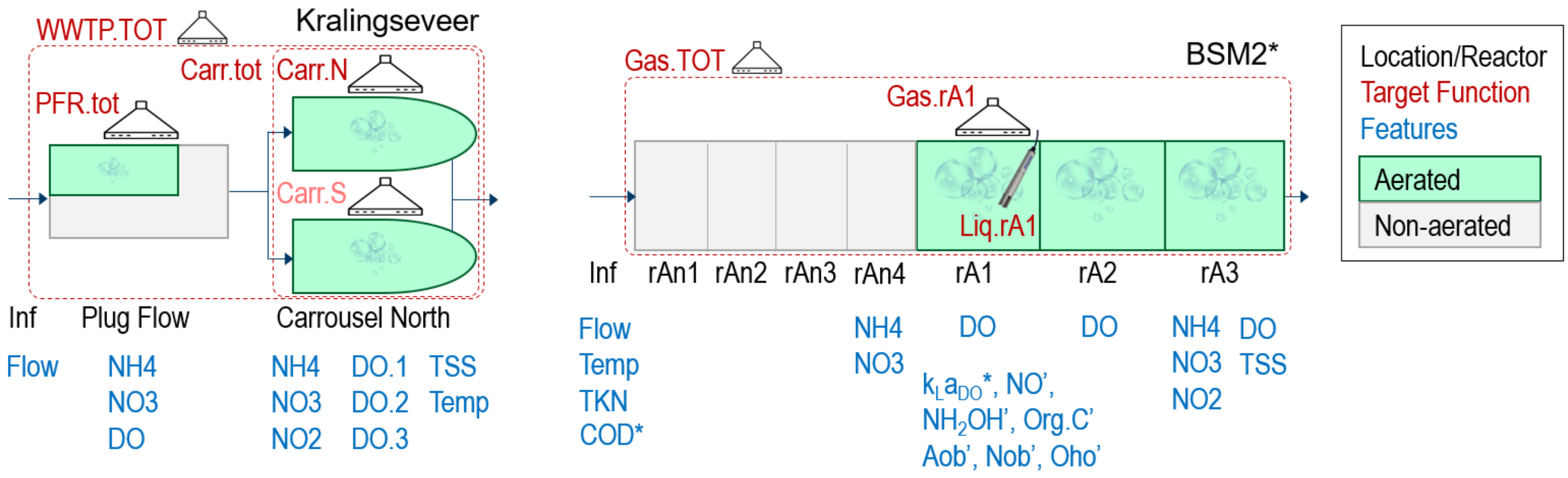

**Figure 1**. Diagrams of the monitored parts of the WWTPs for real (left) and synthetic (right) datasets. Monitoring type and locations for the target function (red) and features (blue). * and ' correspond to features included in the Scenarios containing 14 and 21 features, respectively.

**Table 1** – Experimental design for the real and synthetic datasets: 229 and 365 days, frequency 1/25 and 1/15 minutes respectively. For details see **Table S2.1**.

| Scenario ID | Simulation (#) | Features | Target function: Location $N_2O$ emissions | Target function | Features | Frequency | Disturbances | Aim and description |
|---|---|---|---|---|---|---|---|---|
| Real data: Kralingseveer* (Daelman *et al.* 2015) | | | | | | | | |
| Kral_Carr_N | - | 12 | Gas: $Carr_N$ | x | | | | Assessment of the modelling methodology against a real dataset. |
| Kral_Carr_tot | - | 12 | Gas: $Carr_N$+$Carr_S$ | x | | | | Assess different target functions: Sum from parallel Carrousels North and South. |
| Kral_WWTP_tot | - | 12 | Gas: WWTP | x | | | | Assess different target functions: Sum of all emissions (Carrousels and Plug Flow). |
| Kral_PF_tot | - | 12 | Gas: PlugFlow | x | | | | Assess different target functions: Plug Flow. |
| Synthetic data: Variability of Target Function ($N_2O$ emission) and Features type and location | | | | | | | | |
| $N_2O$_Gas_rA1 | 1 | 12 | Gas: x1 Reactor | x | x | | | Representative campaign for monitoring $N_2O$ emissions: one reactor rAerobic1, gas. |
| $N_2O$_Gas_TOT | 1 | 12 | Gas: total site | x | | | | Representative campaign for monitoring $N_2O$ emissions: site-level, gas. |
| $N_2O$_Liq_rA1 | 1 | 12 | Liquid: x1 Reactor | x | | | | Representative campaign for monitoring $N_2O$ emissions: one reactor rAerobic1, liquid. |
| $N_2O$_Gas_rA1_14feat | 1 | 14 | Gas: x1 Reactor | | x | | | Additional information: aeration rate one reactor ($k_La_{O2}$), influent C-load (COD). |
| $N_2O$_Gas_rA1_21feat | 1 | 21 | Gas: x1 Reactor | | x | | | Additional information: concentration of soluble organic carbon fractions, $NH_2OH$, NO, aob, nob, and oho. |
| Synthetic data: Variability of WWT scenarios. Features from N2O_Gas_rA1_14feat. | | | | | | | | |
| Baseline | 1 | 14 | Gas: total site | | | x | x | Representative campaign for monitoring $N_2O$ emissions: site-level, gas. |
| Baseline_1h** | 1 | 14 | Gas: total site | | | x | | Assess lower monitoring frequencies (1 / 1h) against operational time scales. |
| Baseline_3h** | 1 | 14 | Gas: total site | | | x | | Assess lower monitoring frequencies (1 / 3h) against operational time scales. |
| Mass Transfer Eq. | 2 | 14 | Gas: total site | | | | x | Different liquid-gas mass-transfer calculation: assumption of partial equilibrium. |
| Aerobic Volume | 3 | 14 | Gas: total site | | | | x | Different operation of reactors due to aerated dead zones: 80% aerobic volume. |
| Microbio. Non-$N_2O$ | 4 | 14 | Gas: total site | | | | x | Different microbial community associated to N-processes, except for $N_2O$: ± 10% in kinetic parameters (n = 26). |
| Microbio. $N_2O$ | 5 | 14 | Gas: total site | | | | x | Different microbial community associated only to $N_2O$ processes: ± 20% in kinetic parameters (n = 17). |
| Influent | 6 | 14 | Gas: total site | | | | x | Different influent characteristics: increased N-load, temperature and rain events. |
| Biased $N_2O$ dataset | 7 | 14 | Gas: total site | | | | x | Unknown error in $N_2O$ data: features from Sim. 1, $N_2O$ emissions from combined biases in Sim. 2 to 6. |
| 6-year Dataset** | 1:6 | 14 | Gas: total site | | | | x | Combination of datasets with different souces of variability: Simulations 1 to 6. |
| Biological structure | 8 | 14 | Gas: total site | | | | x | Different assumptions in the biological production of $N_2O$ by AOB. |

** The original dataset was adapted, see Methods. **The dataset was downsampled to lower frequencies.*

### 2.1.2. Synthetic data generation.

A full-scale model for WWT that includes GHG emissions was used to generate yearly monitoring campaigns for $N_2O$ emissions (n = 16). The modified BSM2 layout simplifies the WWT layout from (Solís et al., 2022) and focuses on the N-removal processes as shown in Fig. 1 (right) and Fig. S2.1. The layout does not intend to mimic the Kralingseveer WWTP and consists of primary clarifier, activated sludge and secondary clarifier, while the solids line and P-removal processes were not included (thickener, anaerobic digestor, dewatering, storage tank). The modified activated sludge reactors operated with anoxic zones (4 x 750 $m^3$) and aerobic zones (3 x 3000 $m^3$). The biokinetic models follow (Solís et al., 2022) with the addition of anoxic gaseous emissions ($k_La_{anoxic}$ = 1.1 $day^{-1}$). The $N_2O$ model structure includes two autotrophic pathways (nitrifier nitrification and denitrification) and a heterotrophic pathway (four-step denitrification) (Flores-Alsina et al., 2025). The liquid and gas emissions from each activated sludge tank were calculated for $N_2O$ and NO emissions, the individual pathway contributions and EF. All the simulations consist of 300 days of steady state followed by 609 days of dynamic influent. The datasets comprise the last 365 days of the dynamic simulations and data acquisition happens every 15 minutes and measurements are instantaneous with no error.

**Scenarios**: Sixteen scenarios were generated from eight simulations of the full-scale model (**Table 1**). From Simulation 1 eight different scenarios were generated, while one scenario was generated from each of the other eight simulations, which mimics a total of sixteen monitoring campaigns. The differences between scenarios are the target function, the features, and disturbances to WWT operation. The aim and description of each scenario is described in **Table 1**, more details in SI-S2.

- **Target function**. Three different $N_2O$ emissions monitoring methodologies were analysed: from the first aerated reactor gaseous emissions (Gas.rA1, gN/d), liquid $N_2O$ concentration (Liq.rA1, gN/$m^3$), and total gaseous WWTP emissions (Gas.TOT, gN/d). The initial 12 features selected from the modified BSM2 layout aim to mimic those from Kralingseveer WWTP by (Khalil et al., 2023) ($NH_4^+$, $NO_2^-$, $NO_3^-$, DO, TSS, Temperature, TKN, Flow rate) (**Fig. 1**).
- **Feature number and type**. We investigated whether potential sensors for the substrates and catalysers of $N_2O$ production pathways were more informative than current TN-driven monitoring sensors such those in the Kralingseveer dataset. The target function was the gaseous emissions from the first aerated reactor (Gas.rA1). The availability of two and nine more sensors in the first aerobic tank (rA1) was considered for a total of: 14 features (mass transfer coefficient for aeration, concentration of COD in the influent), and 21 features (hydroxylamine ($NH_2OH$), and nitric oxide (NO), soluble organic carbon, total abundance of heterotrophic bacteria, ammonia oxidisers and nitrite oxidisers.
- **WWT disturbance**. For the same gaseous emission from the total WWTP (Gas.TOT), eight different scenarios were simulated: (#2) not assuming equilibrium in the $N_2O$ liquid-gas mass transfer rate, (#3) the volume of the aerobic reactors reduced 20% to mimic poor mixing zones, (#4) the biokinetic parameters not associated to $N_2O$ production (± 10%), (#5) the biokinetic parameters associated to $N_2O$ production (± 20%), (#6) the influent characteristics (higher N-load per person, infiltration and rain, and mean temperature), (#7) the data quality, (#1:6) where the previous disturbances combined into a 6-year long dataset but down sampled six times to reflect the simultaneous disturbance effect, and (#8) a different biokinetic model structure for $N_2O$ production.
- **Data acquisition frequency**: The scenario Baseline (14 features) was down sampled from the original frequency of 15 minutes to 1 hour and 3 hours, generating two additional scenarios. Only one of the four and twelve possible datasets was considered. The target function was site-level gaseous emissions (Gas.TOT).

### 2.2. Models for $N_2O$ emissions.

Linear regression models were developed to predict $N_2O$ emissions with 12 features as in (Khalil et al., 2023), from first order ($n \leq 13$) and second order with interactions ($n \leq 91$). The method of backward stepwise elimination distinguished statistically significant coefficients.

Five ML models previously used to predict $N_2O$ emissions were evaluated: the ensemble models based on decision trees AdaBoost, XGBoost and Random Forest (RF), Support Vector Regression (SVR), and k-Nearest Neighbours (kNN) (Khalil et al., 2023; Szeląg et al., 2023). These models provide a balance between achieving high performance and maintaining a level of interpretability. For easier comparison, datasets were normalized prior to model training and regardless of the size the data was split into subsets using a k-fold cross-validation approach (CV = 5) (Kohavi, 1995). Hyperparameter were optimized with the hyperband algorithm designed to efficiently allocate resources for tuning machine learning models (Li et al., 2018). The list of ranges and optimized values, processing time, and accuracy for Simulation 1, Scenario $N_2O$_Gas_rA1 are detailed in SI-S3. The ML models were implemented in Python 3.12.12 (libraries: numpy, pandas, matplotlib, seaborn, sklearn, statsmodels) and synthetic data generated in a Matlab/Simulink implementation (MATLAB R2024b).

### 2.2.1. Analysis of model performance and $N_2O$ emission datasets.

Trends in $N_2O$ emission datasets were analysed via Pearson correlation coefficients, the dynamics via the Sum of Absolute Differences (SAD, $gN/(m^3 \cdot d)$), and the approximated curvature as the sum of squared second differences (SSD, $gN^2/(m^3 \cdot d^2)$). While SAD measures how much the signal changes between consecutive points, SSD measured the changes of changes or how curvy is the signal. SAD and SSD were normalised to the corresponding aerated volume and monitoring duration (SADn, SSDn). Unlike statistical normalization This physical normalization allows the units to remain physically interpretable.

The performance of the models was evaluated with the coefficient of determination ($R^2$) and model errors with the mean squared error and mean absolute error. Residual were analysed for temporal patterns. To avoid model overfit the train and validation performances were compared, ensuring lower validation performances. Permutation feature importance was used to determine the significance of input features in the ML models, evaluated with $R^2$. The number of ranked features accounting for over 90% of the model accuracy was considered representative for good fit. The performance of the linear and second order models was compared with the adjusted $R^2$.

## 3. Results & Discussion

### 3.1. Modelling $N_2O$ emissions from the Kralingseveer WWTP.

#### 3.1.1. $N_2O$ emissions in the Kralingseveer WWTP.

The reduced dataset from the Kralingseveer WWTP contains 12 features and 4 target functions (n = 13182). In our reduced dataset mean total $N_2O$ emissions were 94.5 kgN/d, with the Carrousel North, Carrousel South, and the mean of PFRs contributing to 38.9, 30.9 and 24.7 kgN/d respectively, and both Carrousel reactors contributing to 74% of the total (**Fig. 2**, **Table 2**). Additionally, emissions from Carrousel North ($Carr_N$) had very similar patterns to those from both Carrousels ($Carr_{tot}$) and total ($WWTP_{tot}$), with correlations of 0.99 and 0.96 respectively. However, emissions from the PFRs ($PFR_{tot}$) did not correlate to neither $WWTP_{tot}$ (0.20) nor $Carr_{tot}$ (-0.07).

The dataset for total emissions, $WWTP_{tot}$, buffers the dynamic contributions from the three measuring locations with a lower cumulative variability (SADn 14.0 gN/(m$^3$·d)) compared to the Carrousels (21.1, 17.0 gN/(m$^3$·d)) and mean PFR (17.4 gN/(m$^3$·d)) (**Table 2**). Also, the approximated curvature (SSDn) of $WWTP_{tot}$ is lower than that in the Carrousels but higher than in PFR, which has the lowest skewness and curvature of the four $N_2O$ emission datasets. Hence, the oscillatory behaviour in the Carrousels is damped as the $N_2O$ flows are aggregated, i.e. different peak frequencies and amplitudes of $N_2O$ emissions. Overall, the Kralingseveer dataset permits the comparison of alternative monitoring campaigns that capture different spatial and temporal dynamics. Diurnal and operational patterns are observed in high autocorrelation values, for example, 0.98 for consecutive data points decreasing to 0.87 and 0.53 in 2.5 and 10 hours, and peaking daily. For detailed description of the dataset see SI-S1.

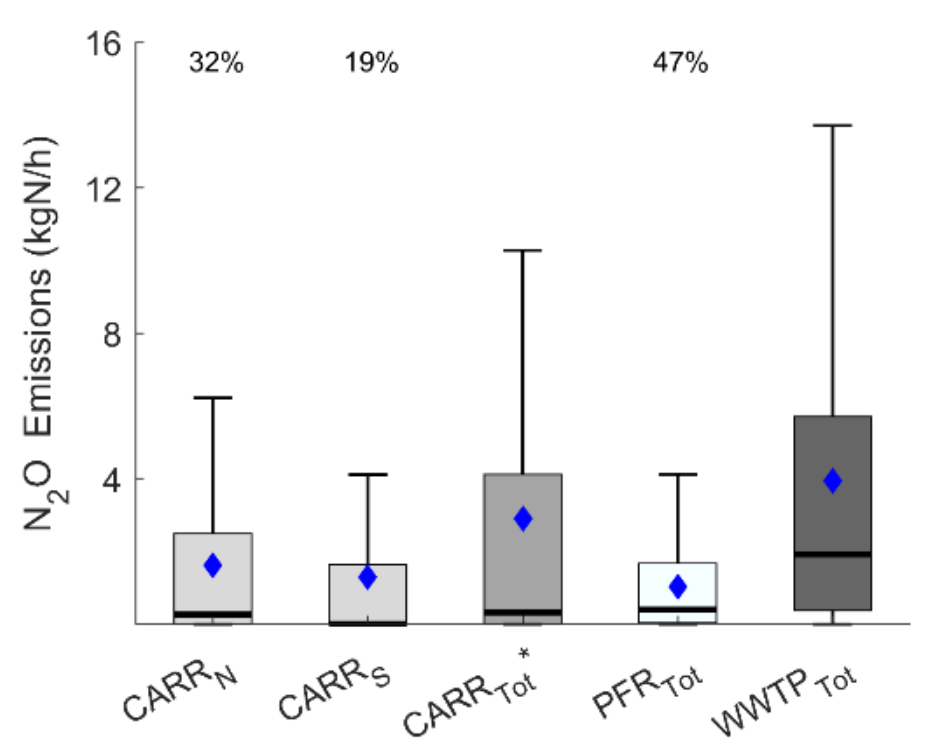

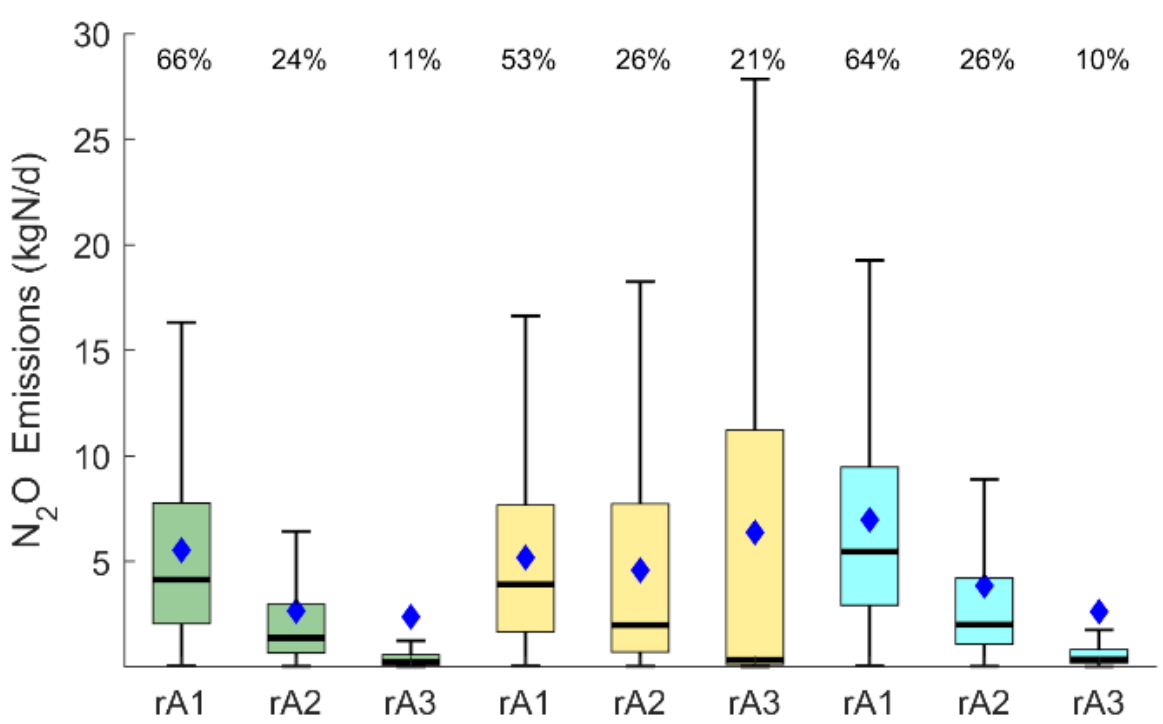

**Figure 2**. $N_2O$ emissions for the target functions from the real dataset (left) and three synthetic datasets (Simulations 1, 3, 7: Baseline green, Aerobic Volume gold, Biased $N_2O$ Dataset cyan). The values above the boxplot represent the average contribution (%) to the site-scale emissions from the WWTP during the monitoring period.

**Table 2** – Distribution of spatially variable $N_2O$ emissions from real and synthetic datasets (Simulation 1, 3, and 7, Table 1).

| | | Real dataset Kralingseveer | | | | Synthetic dataset Baseline | | | | Aerobic Volume | | | | Biased $N_2O$ dataset | | | |
|---|---|---|---|---|---|---|---|---|---|---|---|---|---|---|---|---|---|
| | | $WWTP_{TOT}$ | $Carr_N$ | $Carr_{tot}$ | PFR | TOT | rA1 | rA2 | rA3 | TOT | rA1 | rA2 | rA3 | TOT | rA1 | rA2 | rA3 |
| **Mean** | (kgN/d) | 94.5 | 38.9 | 69.8 | 24.7 | 10.5 | 5.5 | 2.6 | 2.4 | 16.1 | 5.2 | 4.6 | 6.4 | 13.5 | 7.0 | 3.9 | 2.6 |
| **Stdv** | (kgN/d) | 114.6 | 58.3 | 112.5 | 31.2 | 11.4 | 4.5 | 3.3 | 5.7 | 17.3 | 4.3 | 5.2 | 9.9 | 13.7 | 5.5 | 4.8 | 6.4 |
| **CV** | (%) | 121 | 150 | 161 | 126 | 108 | 82 | 123 | 243 | 107 | 84 | 114 | 155 | 102 | 80 | 124 | 245 |
| **SADn** | $gN/(m^3 \cdot d)$ | 14.0 | 21.1 | 17.0 | 17.4 | 9.7 | 13.7 | 7.9 | 8.9 | 12.1 | 13.5 | 10.7 | 14.6 | 12.6 | 16.5 | 12.6 | 10.7 |
| **SSDn** | $gN^2/(m^3 \cdot d^2)$ | 8.4E+5 | 1.2E+6 | 1.1E+6 | 1.7E+5 | 2228 | 1541 | 1069 | 2266 | 4090 | 3136 | 1736 | 4625 | 3681 | 1953 | 2824 | 2927 |
| **Skewness** | ( - ) | 1.52 | 1.72 | 1.82 | 1.29 | 2.12 | 1.28 | 2.16 | 3.04 | 1.44 | 1.12 | 1.32 | 1.50 | 2.09 | 1.52 | 2.34 | 3.32 |
| **Correlation to site-scale** | | 1 | 0.96 | 0.96 | 0.20 | 1 | 0.71 | 0.96 | 0.88 | 1 | 0.63 | 0.98 | 0.95 | 1 | 0.70 | 0.95 | 0.84 |
| **$Mean_{Volume}$** | $gN/(m^3 \cdot d)$ | 2.7 | 2.8 | 2.5 | 3.1 | 1.2 | 1.8 | 0.9 | 0.8 | 1.8 | 1.7 | 1.5 | 2.1 | 1.5 | 2.3 | 1.3 | 0.9 |

#### 3.1.2. Predicting $N_2O$ emissions from Kralingseveer WWTP: Carrousel North.

To compare our methodology against previous studies the first target function analysed was $Carr_N$ (Khalil et al., 2024, 2023; Vasilaki et al., 2020, 2018). Importantly, the true site-level $N_2O$ emissions from Kralingseveer are unknown, and 94.5 kgN/d are a lower bound estimate of unknown uncertainty. Hence, emissions from $Carr_N$ represent up to 41% of the emissions from this monitoring campaign, not the site-level emissions. Four of the five ML models predicted the dataset of $N_2O$ emissions accurately for the four scenarios from Kralingseveer WWTP ($R^2 > 0.8$ except SVR, 0.79 – 0.89) (**Fig. 3**). For the $Carr_N$ scenario the most influential features across ML models were $NH_4^+{}_{PFR}$, $NO_2^-{}_{CARR}$, $NO_3^-{}_{CARR}$, and $TEMP_{CARR}$, but the top ranking differed between ML models (**Fig. 3**). The feature importances across ML models only correlated between AdaBoost and RF (0.93) and was low for the rest of ML models (0.24 ± 0.18). Also, the number of features describing 90% of the feature importance was 9 (RF, kNN), 10 (AdaBoost) and 11 (XGBoost) (SI-S4). Based on these results it is difficult to generalize the most influential features because every ML model might predict $N_2O$ emissions accurately with alternative sensors or sensor locations. One reason could be due to the nature of ML algorithms themselves. Tree-based ensemble models provide piecewise constant predictions with step-line patterns (Micallef et al., 2025), while kNN is affected negatively by noisy or irrelevant features (Halder et al., 2024).

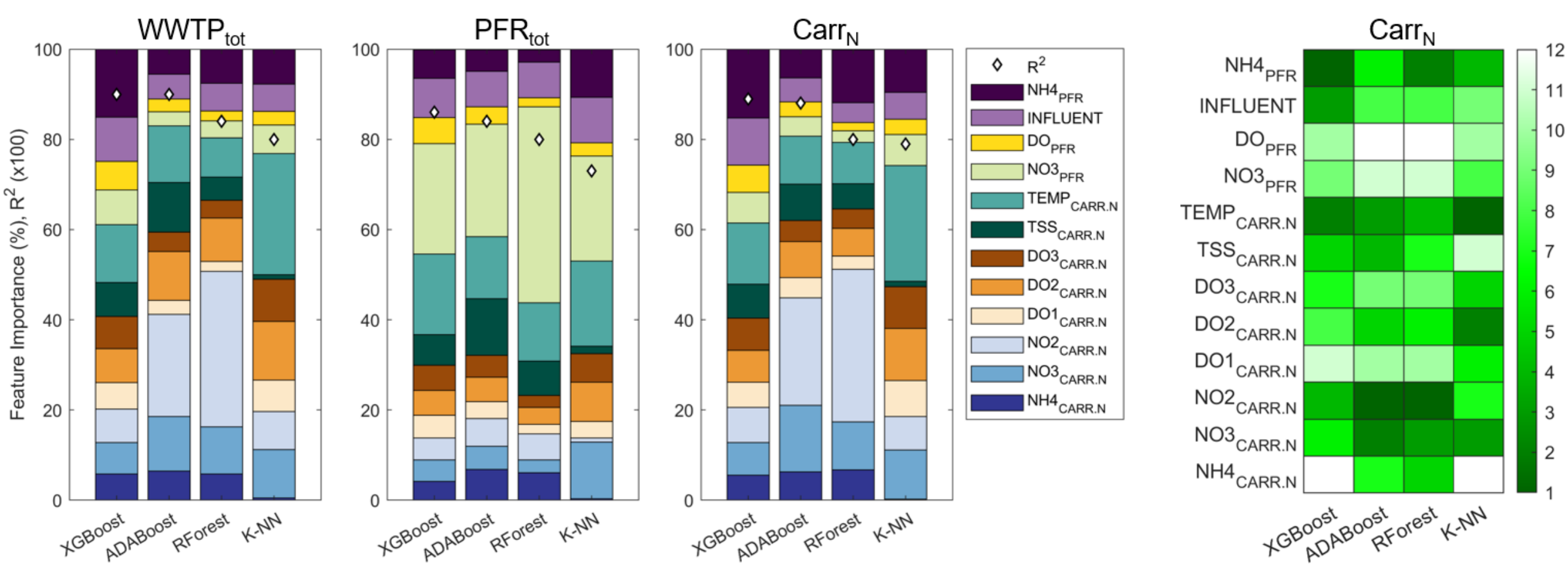

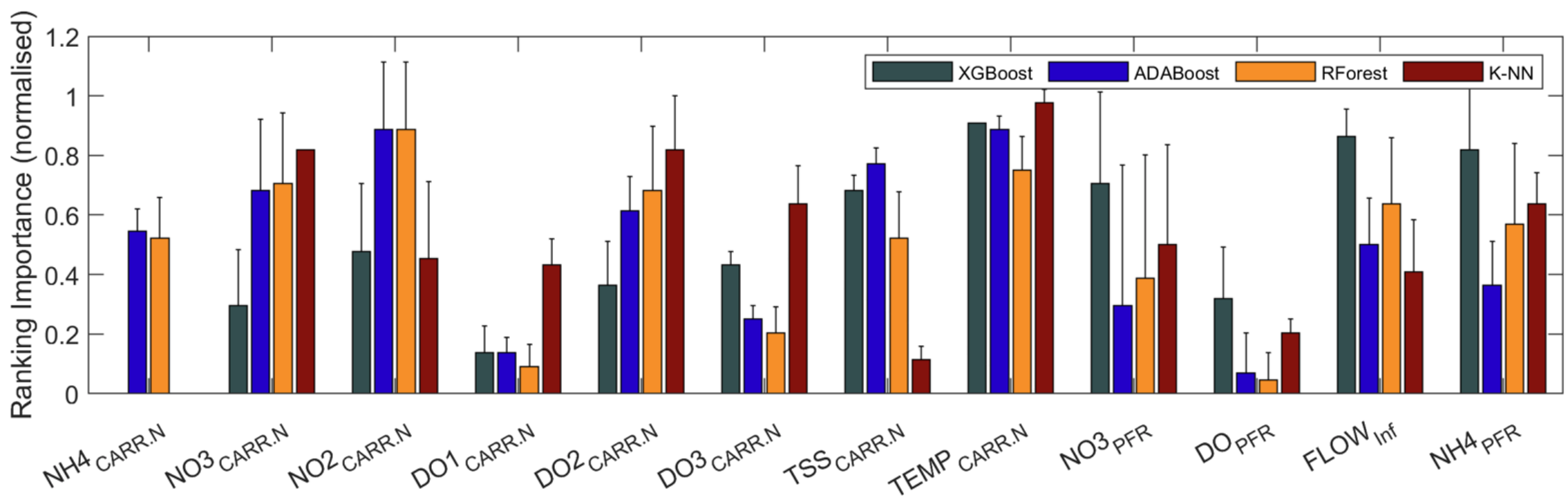

**Figure 3**. Top: Kralingseveer WWTP, accuracy ($R^2$, marker) and feature importance (bar) for three target functions: Total emissions ($WWTP_{tot}$, left), Plug Flow Reactor ($PFR_{tot}$, middle), Carrousel North ($Carr_N$, right). Heatmap for the ranking of feature importances ($Carr_N$), lower values correspond to higher influence. Bottom: Average normalised ranking for the feature importances of the four target functions ($Carr_N$, $Carr_{tot}$, $PFR_{tot}$, $WWTP_{tot}$), higher values correspond to higher influence.

The accuracy for four ML models was approximately 10% lower than other ML models for the same target function, $Carr_N$ ($R^2$ = 0.83 – 0.95) (Khalil et al., 2023). Neither the raw dataset (13 *vs.* 12 features), data pre-processing or hyperparameter optimization procedures were the same, but for AdaBoost, the feature importance correlated highly to two alternative methodologies (0.94 and 0.95) (Khalil et al., 2024, 2023). The aim of this study was not to develop a modelling framework and thus, we assumed that the current methodology was adequate for further analysis of additional scenarios with a focus on the most influential features.

### 3.1.3. Predicting the spatial variability of $N_2O$ emissions in Kralingseveer WWTP.

We analysed how alternative designs of monitoring campaigns (i.e. different target functions) will impact ML model predictions of $N_2O$ emissions. We assumed that $WWTP_{tot}$ is the most informative target function available, while $Carr_{tot}$ and $PFR_{tot}$ are biased spatially because they capture different fractions of the total emissions as was shown in Fig.2. All reactors are covered, and $N_2O$ emissions from the two sampling points in the plug flow reactor were very similar and correlated (25.0 ± 31.5, 24.4 ± 31.1 kgN/d, 0.99). The same four ML models were evaluated against $WWTP_{tot}$, $Carr_{tot}$, $PFR_{tot}$, and the accuracy was similar to the results from $Carr_N$ (0.73 – 0.90, 0.84 ± 0.05) (**Fig. 3**). While some features were uninfluential across target functions ($DO_{PFR}$, $DO_{CARR}$) others differed significantly depending on the ML model ($NH4_{CARR}$) (**Fig. 3**). Similar variability in feature influence was observed in a water treatment case study across tree-based, kernel-based, and connectionist models (Panigrahi et al., 2025).

The feature importances for the $WWTP_{tot}$ target function correlate well with those from $Carr_N$: XGBoost 0.64, AdaBoost 0.82, RF 0.90, kNN 0.99 (SI-Table S4.1). The high correlation between $WWTP_{tot}$ and $Carr_N$ (0.96) explains the similarities in feature importance as compared to $WWTP_{tot}$ and $PFR_{tot}$ where $N_2O$ emissions did not correlate (0.20) (**Table 2**): XGBoost 0.49, AdaBoost 0.00, RF -0.13, kNN 0.57 (SI-Table S4.1). The Kralingseveer WWTP benefits from gas collection systems from covered reactors, which solves possible spatial variabilities within the Carrousels and PFR compared to single hood estimations (~1m$^2$). Despite the reactor-scale measurements, the location of $N_2O$ measurements impacted ML model results due to different $N_2O$ net emissions and dynamics within the WWTP (**Table 2**).

The number of accurate ($R^2 > 0.9$) ML model predictions for $N_2O$ emissions in literature is increasing rapidly. These models have been developed using datasets collected with a wide range of sensor types and spatial sampling scales (Hwangbo et al., 2021; Seshan et al., 2025; Szeląg et al., 2023). For example, a dataset where gaseous $N_2O$ concentrations are monitored in one of the seven covered lanes (Seshan et al., 2025) is more representative of site-level emissions than one of four open parallel reactors monitored with liquid $N_2O$ sensors (Hwangbo et al., 2020), and more than liquid and off-gas measurements in two locations of the open aerated zones during a 96-hour period due to the short duration (Szeląg et al., 2023). Despite these differences in representativeness, developing accurate soft sensors allow for real liquid or gas $N_2O$ sensors to be relocated and help elucidate the spatial variability.

To compare the performance of models of varying complexity, a linear model was estimated for $Carr_N$ yielding low accuracy ($R^2_{adj} = 0.46$, n = 12) compared to a linear model with first-order interactions ($R^2_{adj} = 0.66$, n = 68). Neither improved significantly after aligning for cross-correlation to the $N_2O$ dataset (see SI-S3.1). This suggests that the underlying relationships are likely nonlinear, multi-factorial, and interaction-heavy, beyond what polynomial expansions can meaningfully capture. Alternative regression and self-correlated models could provide good performance for nowcasting (Vanrolleghem et al., 2025). For example, a seasonal autoregressive integrated moving average model with exogenous variables predicted succesfully primary effluent $NH_4^+$ and COD (RMSE < 15%) (Haimi et al., 2025). Here, the autocorrelation structure revealed strong temporal persistence and diurnal variation in $N_2O$ emissions, which are the result of both short-term process memory and daily operational cycles (Fig. S1.3). Hence, alternative simpler models could be considered in the future to balance the bias-variance tradeoff and prevent model accuracies higher than the analytical uncertainties of the $N_2O$ datasets (10 to 25%) (Bollon et al., 2016; Domingo-Félez et al., 2024; Myers et al., 2021).

### 3.2. Nitrogen removal and $N_2O$ emissions in the modified BSM2 layout.

To overcome the limitation of existing datasets designed to control TN instead of $N_2O$ emissions we then used synthetic datasets as a diagnostic tool to analyse their information content (**Table 1**). From the 18 one-year long $N_2O$ monitoring campaigns 8 datasets originated from Simulation 1, the Baseline scenario. The N-removal in Simulation 1 was representative of low nitrogenous effluent concentrations ($NH_4^+ = 0.2 \pm 0.5$ gN/m$^3$, $NO_3^- = 5.1 \pm 1.3$ gN/m$^3$) and an $N_2O$ emission factor of 1.09 ± 1.11% for WWTP emissions, which captures the high temporal variability. The average contributions along the aerated reactors rA1, rA2 and rA3 were 64%, 24% and 12%, and the site-level emissions (Gas.TOT) correlated more to emissions from rA2 (0.96) than from rA1 or rA3 (0.74, 0.82) (**Fig. 2**). Hence, while the first aerobic zone emitted the most $N_2O$, the second aerobic zone was more representative of the site-level emission dynamics. Similarly, while $N_2O$ liquid concentrations correlate well to gaseous emissions from the aerated reactors rA1, rA2 and rA3 (0.89, 0.96, 0.99), rA2 correlates the most to total emissions Gas.TOT compared to rA1 and rA3 (0.68, 0.92, 0.86) (SI-Fig. S2.2). Diurnal and operational patterns are observed in high autocorrelation values, for example, 0.99 for consecutive data points decreasing to 0.86 and 0.55 in 1 and 2 hours (SI-Fig. S2.3). However, the autocorrelation patterns are different to the real dataset, with no smooth decay and more oscillatory signal. The largest contribution to $N_2O$ emissions from the three aerated reactors was the nitrifier denitrification pathway (93, 81, 79%), followed by the heterotrophic denitrification (5, 7, 13%) and nitrifier nitrification pathways (2, 12, 8%). Liquid emissions to the secondary settling tanks were in average 3% of total emissions and included as site emissions. See SI-S2 for more details on spatial and temporal variations.

In Simulations 2 to 8 the N-removal performance did not change substantially except for Scenario Aerobic Volume, where effluent $NH_4^+$ increased to 0.62 ± 1.28 gN/m$^3$, $NO_3^-$ decreased to 4.22 ± 1.59 gN/m$^3$, and the EF was higher at 1.75 ± 2.22% (SI-Table S2.2). The emissions from Kralingseveer are larger than those from the modified BSM2 layout (2.7 *vs.* 1.2 ± 0.3 gN/(m$^3$·d)), but overall, the synthetic datasets are in range of full-scale emissions (**Table 2**) (Gruber et al., 2021). The average EF across the disturbances varied between 0.91 and 1.45% and the correlations between the target functions were higher than 0.87 except for Scenarios Aerobic Volume and Biological structure (0.28 - 0.65) (SI-Fig. S2.2).

Overall, the net $N_2O$ production rate and the pathway contributions vary both temporally and spatially with the environmental conditions simulated (SI-Fig. S2.4). $N_2O$ dynamics were highest in the first reactor for all disturbances except Aerobic Volume (SADn, SSDn, **Table 2,** SI-Fig. S2.6). Also, the last aerated reactor was the most positively skewed: very frequent low emissions and rare high emissions. Interestingly, comparing Baseline and Aerobic Volume shows that if only rA1 was monitored the total emissions would only differ by 6% and would not capture the true site-scale difference in emissions of 53% (**Fig. 2**, **Table 2**). The emission patterns in rA3 change significantly, with much larger dynamics and curvature which would not be captured by only monitoring one-reactor. Hence, for a mitigation strategy to be validated the spatial variability along the water treatment must be monitored.

### 3.3. Prediction of $N_2O$ emissions from synthetic datasets.

The ML models were evaluated against the synthetic datasets from ideal sensors, and the accuracy of the models was higher than for the datasets from Kralingseveer ($R^2$ = 0.97 ± 0.02, n = 80). XGBoost showed the highest overall accuracy ($R^2$ = 0.99 ± 0.01, n = 16), followed by AdaBoost, Random Forest, kNN, and SVR the lowest (0.95 ± 0.04, n = 16). The differences between ML models were negligible and only one dataset was much lower than the rest for Random Forest and SVR (Liq.rA1) (0.83, 0.93) (**Fig. 4**). Overall, all the ML models predicted the trends of $N_2O$ emissions from different campaigns regardless of the type of $N_2O$ measurements, the number of features, the frequency of data, or the simulation of disturbances to the Baseline scenario. Additional performance metrics for the ML models such as residual analysis, optimized hyperparameters, computing time, convergence plots and error analysis are reported in Supporting Information (SI-S3).

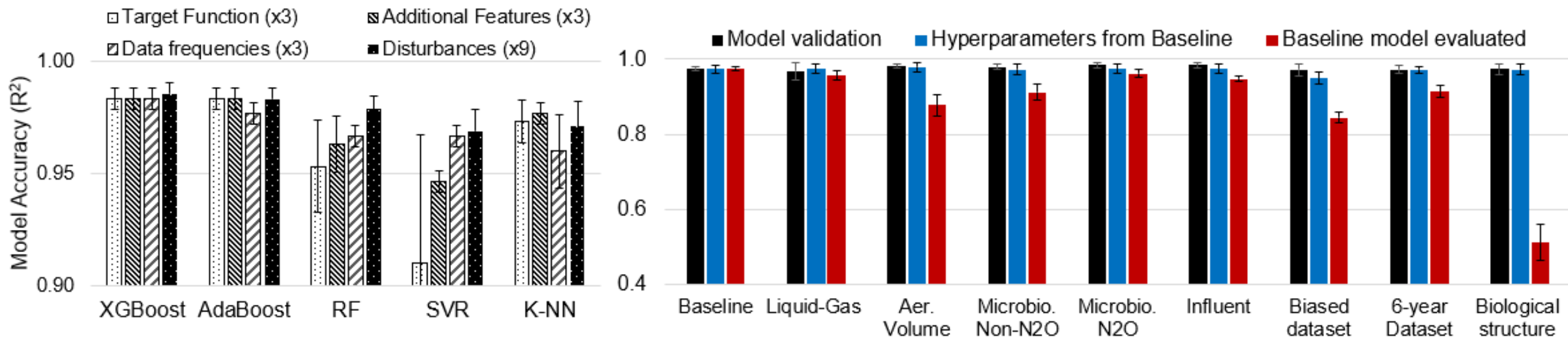

**Figure 4**. Left: Accuracy of ML models evaluated against the synthetic datasets grouped by objectives. In parenthesis number of scenarios included (See Table 1). Right: Performance of ML models for Disturbances with validated models with optimized hyperparameters (black), validated models with the optimized hyperparameters from Baseline (blue), and evaluation of the model Baseline (Scenario 1) for the rest of the scenarios (Scenario 2:8).

For example, in the dataset with 12 features where gaseous $N_2O$ emissions are monitored in one reactor (Gas.rA1, **Fig. 5**), the top ranked feature varied between $NO_3^-{}_{rA3}$ (XGBoost, AdaBoost, RF) and $NH_4^+{}_{rA1}$ (SVR, kNN). The importance ranking of features between ML models did not show patterns for any of the 16 scenarios, and the number of features required to predict 90% of the accuracy also varied and averaged between 4 (RF) and 12 (XGBoost) for all 16 scenarios (SI-Fig. S4.1). In other words, ML models do not sacrifice accuracy for consistent, stable feature rankings. Feature ranking depends heavily on model choice and scenario, and this may impose limitations on interpretability. How users set up ML models was suggested to impact interpretations of dependencies between outputs and inputs (Panigrahi et al., 2025). Calibration of the models for Simulation 2 to 9 with the optimized hyperparameters of Simulation 1 did not affect the model accuracy significantly (mean accuracy loss of 0.69 ± 0.72%, n = 40) (**Fig. 4**, blue bars). The ML models trained from the Simulation 1 were also evaluated against unobserved datasets from Simulation 2 to 9 (**Fig. 4**, red bars). The accuracy decreased for the scenarios of Aerobic Volume (11%), Biased $N_2O$ dataset (13%), and the alternative Biological structure (47%). These results represent the sensitivity of ML models trained with one-year datasets to expected disturbances during WWT operation. Biased data from sensor errors, poor mixing zones, changes in microbial communities and mitigation strategies can yield unobserved patterns in $N_2O$ emissions for the ML models. The largest accuracy difference between two $N_2O$ model structures for ammonia-oxidizing bacteria (AOB) highlights the need for a congruent structure across models (Ding et al., 2017; Domingo-Félez and Smets, 2016; Pocquet et al., 2016).

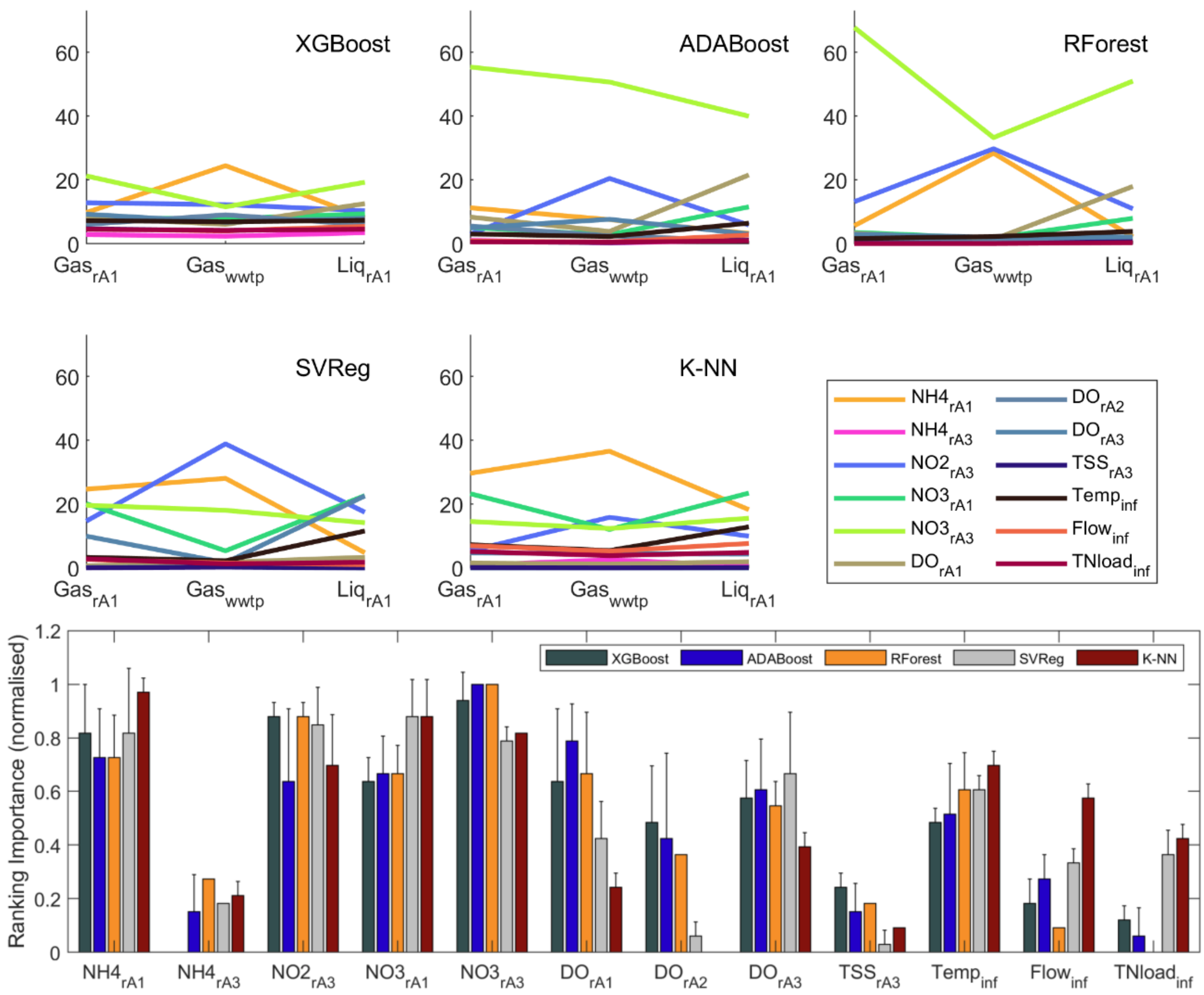

**Figure 5**. Top: Feature importance for ML models trained with synthetic datasets: scenario Baseline, target functions Gas.rA1, Gas.TOT, Liq.rA1. Bottom: Average normalised ranking for the feature importances of three target functions, higher values correspond to higher influence.

### 3.3.1. Effect of Target function: $N_2O$ measuring technology and location.

ML models were calibrated with synthetic datasets for three different target functions capturing emissions from one aerated reactor, gaseous (Gas.rA1) and liquid-based (Liq.rA1), and total WWTP emissions (Gas.TOT) (**Table 1**). The target function selected affected the importances and rankings differently for each Algorithm (**Fig. 5**, SI-S4). Results from Gas.TOT showed different feature patterns than a single monitoring location and gaseous emissions correlated more with liquid-based measurements (0.74 ± 0.15 vs. 0.57 ± 0.19, n = 5). The feature importance for $NH_4^+$ and $NO_3^-$ sensors show that features from rA1 were more informative than in rA3 for SVR and kNN, as compared to XGBoost, AdaBoost and RF, indicating that both sensor location and ML model impact feature importance. Overall, ML models trained on single-location were not representative for site-level emission patterns and soft sensors developed on single-location will not validate site-level mitigation strategies.

### 3.3.2. Effect of Additional Features.

To capture the influence of alternative sensors, $N_2O$ emissions were predicted from two datasets that include: two common measurements in WWT such as influent COD and aeration rates, and seven measurements for the substrates and catalysts of $N_2O$ emissions (**Table 1**). The aeration rate represents an accurate estimator for the $N_2O$ stripping rate, and $k_La_{DO}.rA1$ ranked between 1$^{st}$ and 6$^{th}$ in feature importance across all models (SI-Fig. S4.5). Contrary to our hypothesis, the influent COD, and therefore C/N, did not impact $N_2O$ prediction and ranked, in average, 10$^{th}$. The addition of 9 measurements reduced the importance from the previous 12 features by 49%. The potential measurement of a substrate for $N_2O$ production like $NH_2OH$ contributed significantly to XGBoost, AdaBoost and RF, and overall more than the direct substrate for all the pathways, NO. Interestingly, the abundance of AOB only contributed to 2.0% of the prediction accuracy. The larger time scales of microbial growth might not resolve the more dynamic $N_2O$ production rates, as abundance-based methods do not necessarily correlate to activity. However, advances in molecular methods aim to connect dynamics of microbial abundances to growth kinetics, which help predict process rates (Cheng et al., 2025).

### 3.3.3. Effect of Data Acquisition Frequency.

To analyse how the frequency of online data availability impacts ML model predictions the 3-hour frequency was selected because of the lower autocorrelation value in the target function (0.21) compared to 1 and 2 hours (0.86 and 0.55) (SI-Fig. S4.6). No significant differences in accuracy or feature importance rankings were observed between frequencies of 15 minutes, 1 hour, or 3 hours. The correlation between feature importances across the three frequencies was very high (> 0.98) for XGBoost, RF, SVR and kNN. Only for AdaBoost the 3-hour dataset correlated less to the 15-minute and 1-hour datasets (< 0.75) (SI-S4).

The data acquisition frequency of 15 minutes exceeds the influent, operational time scales or the lagged hydraulics, which has been incorporated into each feature (Wang et al., 2025). The Kralingseveer dataset includes $N_2O$ monitoring for 4 minutes every 25, allowing for a 4-point spatial resolution with a single instrument ($Carr_N$, $Carr_S$, $PF_1$, $PF_2$) (Daelman et al., 2015). Similarly, monitoring 5 minutes every 30 instead of continuously was proposed as equally informative of the $N_2O$ emission distribution while allowing for multi-point monitoring (Domingo-Félez et al., 2024). Hence, the use of high frequency and autocorrelated data should be justified against the time scales in existing datasets, and increasing spatial resolution is recommended instead.

### 3.3.4. Effect of disturbances during WWT

A set of disturbances from the reference scenario (Baseline) were simulated to analyse the prediction of ML models on a range of WWT conditions. The effect of disturbances did not show a pattern neither in the most influential features nor the ML model considered (**Fig. 6**, SI-Fig. S4.7). The effect of the liquid-gas mass transfer calculation, different influent characteristics, or variability in the model biokinetic parameters did not show significant differences compared to the Baseline (Simulation 2, 4, 5, 6). A bias in the $N_2O$ emission measurement only led to variable importance rankings for SVR (Simulation 7). A decrease in the effective aerobic volume of 20% lead to substantially different patterns in $N_2O$ emissions and top ranked features as compared the Baseline results (Simulation 2, **Fig. 6**). Lastly, the description of the microbial $N_2O$ production pathways for AOB with a different structure impacted predictions across all ML models (Simulation 8).

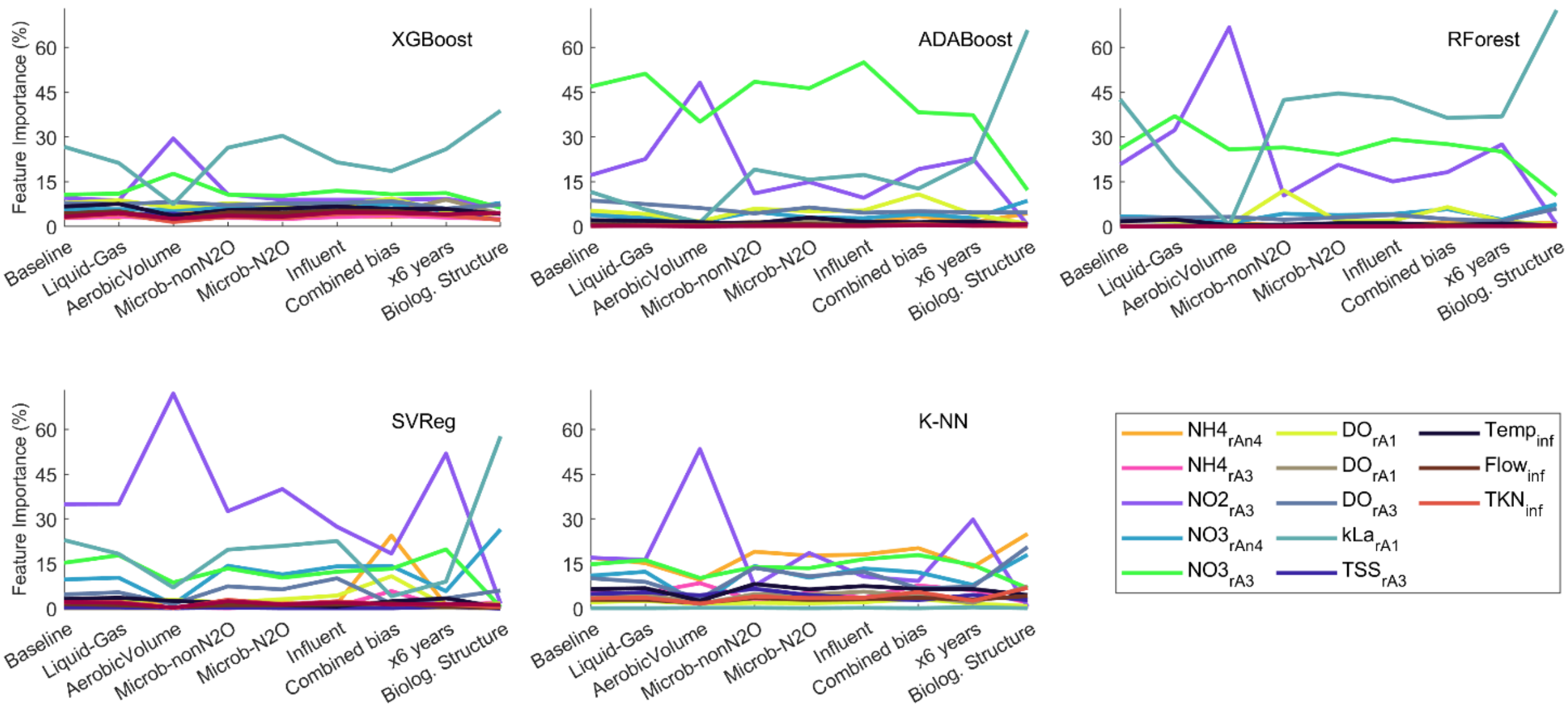

**Figure 6**. Feature importance for ML models trained with synthetic dataset: scenarios Disturbances, target function Gas.TOT.

Accurate soft sensors for $N_2O$ emissions exist for a wide variety of technologies, ML models, and datasets but their limits of interpretability remain understudied often leading to siloed solutions. Soft sensors do not necessarily predict the consequences of mitigation strategies such as changes in microbial communities or the $N_2O$ dynamics at non-monitored locations. When the five ML models were evaluated under unseen synthetic scenarios the top importance features changed, and their performance failed for both significant operational variations and assumptions in microbial reactions (Aerobic Volume, Biological Structure) (**Fig. 4**). One should be cautious about possible model overfit based on the data quality and model robustness when WWT conditions change. Mechanistic models can complement the limitations of soft sensors, site- and data-specific, to forecast $N_2O$ emissions.

### 3.4. The structural robustness of a mechanistic model for $N_2O$ emissions.

We investigated the robustness of a McM to predict spatially variable $N_2O$ emissions. Specifically, how the model structure affects the $N_2O$ production and consumption processes across the secondary treatment and the role of substrates in the production pathways by AOB (NN, ND) and OHO (HD). The inherent interpretability of $N_2O$ production pathways in any McM would help design mitigations for targeted pathways.

**Role of the model structure for $N_2O$ pathways.** The model structure selected is widely adopted in current mitigation exercises (Flores-Alsina et al., 2025; Lei et al., 2025; Li et al., 2025; Solís et al., 2022). The structure of the NN and ND pathways originates from a nitritation reactor with high $NO_2^-$ accumulation and no heterotrophic activity (Pocquet et al., 2016). The stoichiometry of the NN pathway yields $NO_2^-$ anoxically from $NH_2OH$ and NO, which has not been observed from the enzyme hydroxylamine oxidoreductase (Caranto and Lancaster, 2017). The description of oxygen inhibition with a Haldane function in the ND process rate equation was developed originally for a single-pathway model (Guo and Vanrolleghem, 2014), which does not predict the observed anoxic $N_2O$ production in the presence of $NH_2OH$ and $NO_2^-$ (Domingo-Félez et al., 2017a). Hence, during temporal or spatial gradients of DO (e.g. intermittent aeration, deep biofilms) the ND pathway would not contribute to $N_2O$ emissions. The two-pathway model demanded more rigorous evidence to substantiate this DO effect (Pocquet et al., 2016) (SI-S5).

**Role of nitric oxide as biological precursor of $N_2O$ emissions**. Including NO emissions during model calibration benefits elucidating pathway contributions (Pocquet et al., 2016). The estimated NO emissions in Simulations 1 to 7 ranged from 0.33 to 0.55% of total $N_2O$ emissions and the individual NO-to-$N_2O$ values in the aerobic reactors highlight different patterns across the WWTP. The comparison to a different biological $N_2O$ model structure (Simulation 8) indicates a seven-fold difference in NO emissions and different NO-to-$N_2O$, with correlations of 0.57 and 0.93 respectively. Hence, the additional NO emission data would help discriminate between the two model structures. Interestingly, the feature importance of NO liquid dataset in one of the aerobic reactors did not rank high (**Section 3.3.2.**).

**Interactions AOB – OHO**: The model describes aerobic $NO_2^-$ production by AOB as three sequential steps of $NH_4^+$ over $NH_2OH$ and NO. In the first aerated tank the concentration of $NH_4^+$ is highest, and the average process rate of the AMO reaction ($NH_4^+ \rightarrow NH_2OH$) is 25% faster than the HAO reaction ($NH_2OH \rightarrow NO$), as expected from an in-series reaction (AMO > HAO > HAO*) (**Fig. 7**, grey bars). However, the opposite trend occurs for the third step for HAO* reaction ($NO \rightarrow NO_2^-$), and NO production is 6% slower than NO consumption despite also being stripped simultaneously (AMO > HAO < HAO*). The reason for the discrepancy is that AOBs consume heterotrophically-produced NO at a faster rate than the in-series heterotrophic denitrification process, where heterotrophs only consume 5.5% of their NO produced (**Fig. 7**, blue bars). These results are associated to any AOB/OHO model with an NO-loop: $NO_2^-$ is reduced heterotrophically to NO (NIR) and oxidised autotrophically back to $NO_2^-$ (HAO*) (**Fig. 7**, middle). Similar patterns occur for two different values of maximum rate for the HAO step ($Mu_{AOB.HAO}$ = 0.61 and 0.78 $d^{-1}$) and two different influents (Simulation 1 and 6) (SI-S5). Whether AOB can produce more $NO_2^-$ than the own $NH_2OH$ oxidation remains to be elucidated experimentally.

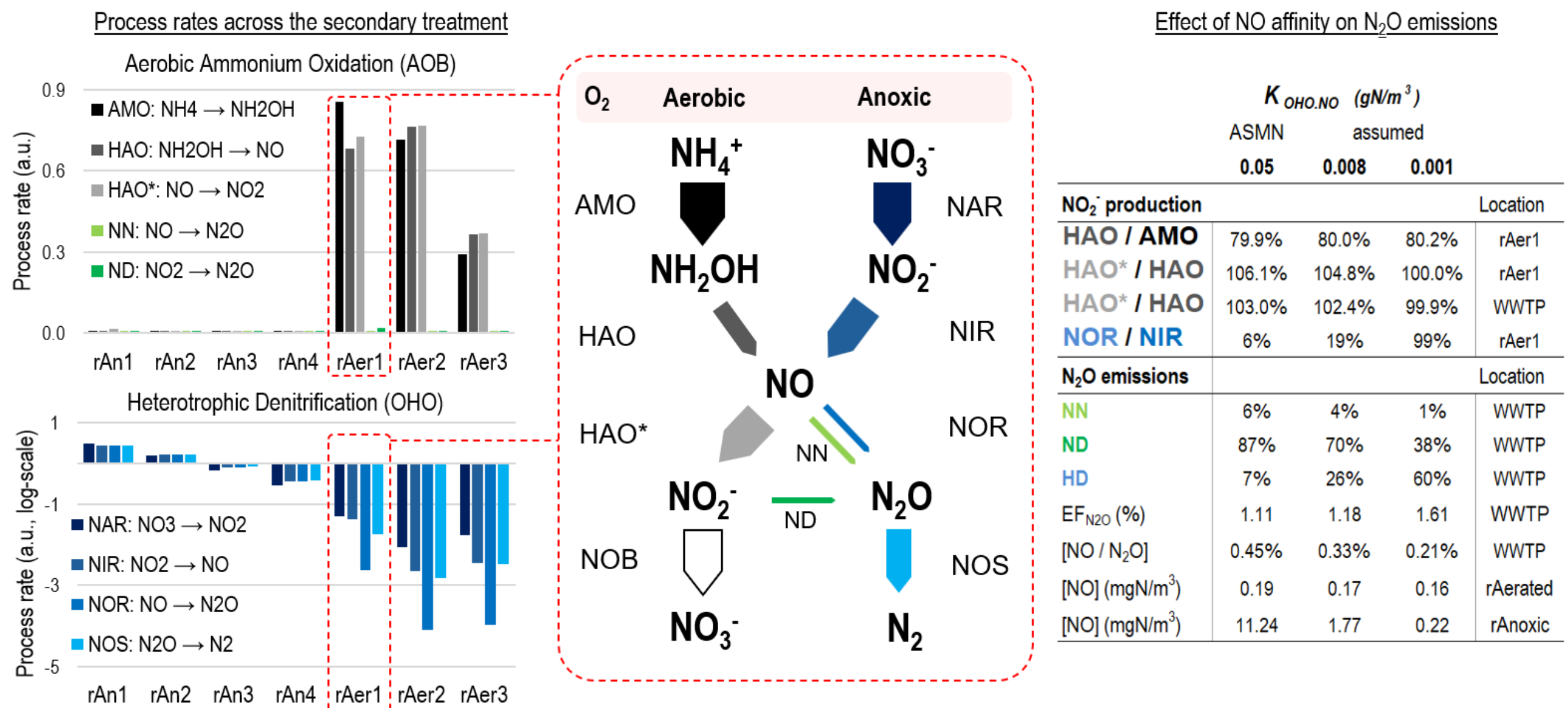

| | $K_{OHO.NO}$ (gN/m³) ASMN | assumed | | |
|---|---|---|---|---|
| | 0.05 | 0.008 | 0.001 | |
| $NO_2^-$ production | | | | Location |
| HAO / AMO | 79.9% | 80.0% | 80.2% | rAer1 |
| HAO* / HAO | 106.1% | 104.8% | 100.0% | rAer1 |
| HAO* / HAO | 103.0% | 102.4% | 99.9% | WWTP |
| NOR / NIR | 6% | 19% | 99% | rAer1 |
| $N_2O$ emissions | | | | Location |
| NN | 6% | 4% | 1% | WWTP |
| ND | 87% | 70% | 38% | WWTP |
| HD | 7% | 26% | 60% | WWTP |
| $EF_{N2O}$ (%) | 1.11 | 1.18 | 1.61 | WWTP |
| [NO / $N_2O$] | 0.45% | 0.33% | 0.21% | WWTP |
| [NO] (mgN/m³) | 0.19 | 0.17 | 0.16 | rAerated |
| [NO] (mgN/m³) | 11.24 | 1.77 | 0.22 | rAnoxic |

**Figure 7**. Left: Average process rates for AOB and OHO across the modified BSM2 layout in Simulation 1. Middle: Schematic of the AOB and OHO interaction in rAer1 (arrows represent rates, not to scale, colour coding as in bar plots). *: unconfirmed. Right: process rates associated to NO and $NO_2^-$ production for varying values of the OHO affinity constants for NO ($K_{OHO.NO}$) for Simulation 1 and $Mu_{AOB.HAO}$ = 0.61 $d^{-1}$. Contribution from each $N_2O$ pathway, Emission Factor, ratio between NO and $N_2O$ emissions, and NO liquid concentrations.

To explain this modelling artifact, we analysed the substrate limitation across the WWTP focusing on NO-processes as the key substrate and product for $NO_2^-$ and $N_2O$ (SI-Fig. S5.1). In the first aerated tank OHO's activity is more NO-limited than AOB's due to a 167-fold larger substrate affinity constant ($K_{OHO.NO}$ = 0.05 gN/m³, $K_{AOB.HAO.NO}$ = 0.0003 gN/m³). If we increase the affinity constant for NO in denitrification processes from the default value of 0.05 to 0.001 gN/m³, the heterotrophic NOR process in rAer1 increases compared to NIR from 6% to 99% (**Fig. 7**, right). Consequently, the third aerobic oxidation step of AOB slows from 106.1% to 100.0% and allows for an in-series aerobic ammonium oxidation process. The increase in heterotrophic affinity ($K_{OHO.NO}$ = 0.001 gN/m³) shifts the contribution of $N_2O$ pathways from autotrophic (87 to 38%, **Fig. 7**, right) towards heterotrophic denitrification (7 to 60%), matching closer to measurements of isotopic measurements of N-species (Gruber et al., 2022; Keck et al., 2026).

The original values for the NO-parameters lack empirical validation, and neither the substrate affinity ($K_{OHO.NO}$) nor the noncompetitive inhibition terms for NIR, NOR and NOS ($K_{OHO.i.NO}$) were traceable (Hiatt and Grady, 2008) (SI-S5). While NO and $N_2O$ emission datasets were not available in 2008, recent efforts on $N_2O$ modelling should question the structural robustness and complexity of ASMN's denitrification process rate equations, especially when coupled to an AOB-only model structure for NN and ND. The compensation between the ND and HD pathways due to their very different values of $K_{NO}$ explains why total emissions are not sensitive to $K_{NO}$ during model calibrations (Domingo-Félez et al., 2017b). If the emissions from individual pathways were considered instead, the key parameters of each pathway would rank highest. While the question of structure compatibility between pathway equations remains unsolved, the predicted results for higher heterotrophic affinity towards NO are more congruent biologically.

## 4. Conclusions

Monitoring campaigns for $N_2O$ emissions include operational data from WWTP currently designed to control TN levels rather than $N_2O$ emissions. Campaigns compromise spatial and temporal resolutions since long-term datasets of site-level emissions are rare. Thus, a meta-analysis of all existing campaigns (Song et al., 2024) could mislead policy makers by including non-representative short campaigns aimed at diagnosing triggers of $N_2O$ production.

- High-accuracy soft sensors based on local $N_2O$ measurements do not represent, unless validated, the site-level EF of the WWTP.
- Four of the five ML models predicted multiple datasets of spatially variable $N_2O$ emissions, but the feature influence differed across models. The inconsistent feature influence limits the general interpretability of the results.
- The methodological uncertainty of $N_2O$ monitoring campaigns tends to be larger than the accuracy of ML models, which raises a concern on possible model overfits.
- Prediction of $N_2O$ emissions from synthetic datasets was more accurate than real datasets. The most influential features depended on the ML model and then on the disturbance predicted. The importance of key features could help prioritize data quality (e.g. sensor maintenance), but depended on the ML model. The synthetic addition of new sensors showed varying importances, from highly relevant for $NH_2OH$ to noninformative for liquid NO. The quantification of microbial abundances did not show a positive influence on the prediction.
- The structure of the McM showed interactions between AOB and OHO during $NH_4^+$ oxidation to $NO_2^-$, with AOBs producing more $NO_2^-$ than its own substrate NO. The role of NO and the substrate affinities ($K_{NO}$) is overlooked and drives the relative contribution of the predicted $N_2O$ production pathways. A higher heterotrophic affinity towards NO than autotrophic yields more congruent results.

Given the high accuracy of current ML models on $N_2O$ point-measurements, future studies should shift the focus towards general limits of interpretability and their robustness.

**Acknowledgements**

We would like to thank Dr. Xavier Flores-Alsina at the Department of Chemical and Biochemical Engineering at the Technical University of Denmark for the support in the modified BSM2 layout implementation from the code available at (www.github.com/wwtmodels).